%% file: acl_latex.tex
\pdfoutput=1

\documentclass[11pt]{article}

\usepackage{acl}

\usepackage{times}
\usepackage{latexsym}
\usepackage{amsmath}
\usepackage{amsfonts}

\usepackage[T1]{fontenc}

\usepackage[utf8]{inputenc}

\usepackage{microtype}

\usepackage{inconsolata}

\usepackage{graphicx}
\usepackage{comment}
\usepackage{booktabs}

\usepackage{array}
\usepackage{caption}

\usepackage{framed}
\usepackage{multirow}
\usepackage{url}
\usepackage{listings}
\usepackage{xcolor}

\usepackage{cuted}

\definecolor{lightyellow}{RGB}{255,255,224}
\definecolor{lightblue}{RGB}{214,237,243}

\lstdefinestyle{metapromptstyle}{
    backgroundcolor = \color{white},  
    basicstyle = \small\ttfamily,           
    breaklines = true,                       
    frame = single,                          
    numbers = none,                          
    captionpos = b,                          
}

\lstdefinestyle{promptstyle}{
    backgroundcolor = \color{white},  
    basicstyle = \small\ttfamily,           
    breaklines = true,                       
    frame = single,                          
    numbers = none,                          
    captionpos = b,                          
}

\title{\ourmethodnosp: Automatic Prompt Induction and Optimization for Grammatical Error Correction and Text Simplification}

\author{
  \textbf{Artem Chernodub\thanks{Accepted for publication at Recent Advances in Natural Language Processing conference (RANLP 2025).}\thanks{The work was done while working at Grammarly.}\thanks{Correspondence: a.chernodub@gmail.com.}\textsuperscript{$\zeta$}} \hspace{5pt}
  \textbf{Aman Saini\textsuperscript{$\gamma$}} \hspace{5pt}
  \textbf{Yejin Huh\textsuperscript{$\gamma$}}
  \hspace{5pt}
  \textbf{Vivek Kulkarni\textsuperscript{$\gamma$}} \hspace{5pt}
  \textbf{Vipul Raheja\textsuperscript{$\gamma$}}
\\
  \textsuperscript{$\zeta$}Zendesk \hspace{5pt}
  \textsuperscript{$\gamma$}Grammarly
}

\begin{document}

\newcommand{\ourmethod}{\textsc{Apio}\space}
\newcommand{\ourmethodnosp}{\textsc{Apio}}
\newcommand{\artem}[1]{{\color{red}\emph{Artem: #1}}}
\newcommand{\aman}[1]{{\color{magenta}\emph{Aman: #1}}}
\newcommand{\vk}[1]{{\color{blue}\emph{Vivek: #1}}}
\newcommand{\vr}[1]{{\color{orange}\emph{Vipul: #1}}}
\newcommand{\yejin}[1]{{\color{teal}\emph{Yejin: #1}}}

\maketitle
\begin{abstract}
Recent advancements in large language models (LLMs) have enabled a wide range of natural language processing (NLP) tasks to be performed through simple prompt-based interactions. Consequently, several approaches have been proposed to engineer prompts that most effectively enable LLMs to perform a given task (e.g., chain-of-thought prompting). In settings with a well-defined metric to optimize model performance, automatic prompt optimization (APO) methods have been developed to refine a seed prompt. Advancing this line of research, we propose \ourmethodnosp, a simple but effective prompt induction and optimization approach for the tasks of Grammatical Error Correction (GEC) and Text Simplification, without relying on manually specified seed prompts. \ourmethod achieves a new state-of-the-art performance for purely LLM-based prompting methods on these tasks. We make our data, code, prompts, and outputs publicly available.\footnote{\url{https://github.com/achernodub/apio}}


\end{abstract}

\input{latex/intro}

\input{latex/apo}

\input{latex/experiment_setup}

\input{latex/results}

\input{latex/related_work}

\section{Conclusion}
We present \ourmethodnosp, a new technique for automatic prompt induction and optimization for the tasks of Grammatical Error Correction and Text Simplification. Our method achieves state-of-the-art performance when compared to other prompting-based baselines on these tasks. \ourmethod represents a significant step forward in automating and simplifying the process of prompt engineering. 

\section{Limitations}
Our research primarily focuses on a limited set of models, and we acknowledge that the design choices—such as the number of prompts, iterations, and other generation hyper-parameters, such as beam size, top-$p$, temperature, etc. have not been exhaustively explored. Additionally, our findings are sensitive to specific model artifacts. We also recognize that we did not investigate other tasks, benchmarks, or languages, which could provide a more comprehensive understanding of the models' effectiveness with respect to APIO.

\section{Acknowledgments}
This research was supported by Grammarly. We are grateful to Mariana Romanyshyn for bringing attention to this topic. We are wholeheartedly grateful to Nastia Osidach, Viktor Zamaruiev, Max Gubin, and Peng Wang for their support. We also thank Shashi Ravula for making it happen. We appreciate the contributions of the three anonymous reviewers.

\bibliography{anthology, latex/custom}

\appendix

\clearpage

\input{latex/appendix_a}

\input{latex/appendix_b}

\input{latex/appendix_c}

\input{latex/appendix_d}

\end{document}

%% file: latex/intro.tex
\section{Introduction}

Prompt engineering has become a popular and crucial technique for steering large language models (LLMs) toward desired outputs, but finding effective prompts remains challenging. Prompting methods like chain-of-thought (CoT) prompting, best-of-n sampling, etc. are general strategies that have been shown to be effective. However, even when these advanced prompting strategies are used, recent studies show that LLMs are highly sensitive to seemingly minor variations in prompts (e.g. phrasing \cite{li2023large}, ordering of information \cite{liu-etal-2024-lost}, or formatting \cite{Sclar0TS24}, which can lead to significant performance variation. Consequently, in practice for many tasks, prompts are tuned by prompt engineers to maximize gains in task performance. Since manual prompt tuning can be tedious, there has been some research on automatic prompt optimization (APO) methods that tune a base prompt based on performance on training and validation sets -- the most relevant work being that of \cite{pryzant-etal-2023-automatic}. 

However, APO, in general, mainly focuses on text classification tasks such as Jailbreak Detection, Math Reasoning, and BIG-bench Hard tasks \cite{zhou2022large, pryzant-etal-2023-automatic, ye-etal-2024-prompt, ma2024large} and has been underexplored for text revision tasks such as Grammatical Error Correction (GEC) and Text Simplification. In this paper, we address this gap and propose a novel prompt induction and optimization method called \ourmethod. In contrast to existing prompt optimization methods that require a seed prompt, \ourmethod does not rely on a manually specified prompt. Instead, it induces a reasonable list of instructions and subsequently optimizes them. In short, \ourmethod performs both automatic prompt induction and optimization. We evaluate \ourmethod\ against strong baselines on standard GEC and Text Simplification benchmarks and show that \ourmethod sets a state-of-the-art performance on these benchmarks. 

Our main contributions are:
\begin{itemize}
    \item We introduce a novel method \textbf{A}utomatic \textbf{P}rompt \textbf{I}nduction and \textbf{O}ptimization (\ourmethodnosp) for text revision tasks (specifically, GEC and Text Simplification).
    \item We set the new state-of-the-art for LLM-based prompting methods on these tasks. For the GEC task, we achieve a score of $59.40$ on the BEA-2019 test dataset, ahead of the previous state-of-the-art ($57.41$) \cite{loem-etal-2023-exploring}. For the Text Simplification task, we achieve a SARI score of $49.47$ on the ASSET-Test dataset, ahead of the previous state-of-the-art ($47.94$) \cite{vadlamannati-sahin-2023-metric}. 
\end{itemize}

%% file: latex/apo.tex
\section{\ourmethod}
\ourmethod has two main steps:

\begin{enumerate}
    \item \textbf{Prompt Induction}. We first induce a prompt given gold-standard examples of task-specific input and output pairs. 
    \item \textbf{Prompt Optimization}. We then optimize the induced prompt to maximize training and validation performance. 
\end{enumerate}


\paragraph{Prompt Induction} Unlike other APO methods which start from an initial, manually crafted seed prompt, \ourmethod requires only a few input–output examples that demonstrate the task —   typically available as training data. Given these examples, we use a state-of-the-art LLM to infer a prompt to solve the task. A key feature of our prompt induction approach is to induce structure to the inferred prompt. In particular, the LLM generates a prompt that consists of a markdown-style list of single-sentence instructions between the prompt's header and footer, which are not optimized (Appendix \ref{sec:appendixA}, Listing \ref{lst:prompt_structure}).

Structuring the prompt as a list of independent instructions allows for instruction-level tuning, and enables more fine-grained control as opposed to tuning a flat text blob. Formally, the output of this step will be a prompt $\mathcal{P}$, consisting of an ordered list of instructions $\mathcal{L}$. Each instruction in the list is derived by the LLM from a single "training" input-output pair (see the meta-prompt for prompt induction in Appendix \ref{sec:appendixB}, Listing \ref{lst:prompt_instruction_induction}). 


\paragraph{Prompt Optimization} In this step, we optimize the induced prompt $\mathcal{P}$ that consists of a list of instructions $\mathcal{L}$ iteratively as follows:

\begin{enumerate}
    \item We consider the instructions in the current pool of size $\mathcal{M}$, which is initialized to $\mathcal{L}$—the set of instructions inferred during the Prompt Induction step.
    \item We then seek to expand the above pool of instructions through a beam search with a beam size $B$. In particular, we expand the pool through three prompting operations:
    \begin{itemize}
        \item \textbf{Improve}: Here, we generate beam candidates by prompting an LLM to improve the given pool of instructions to reduce the error rate on the given input-output examples as much as possible. In our experiments, we use word-level Levenshtein edit distance as a metric for optimization for all domains. See the specific meta-prompt in Appendix \ref{sec:appendixB}, Listing \ref{lst:prompt_improve}. 
        \item \textbf{Rephrase}: Next, we expand the current pool by prompting the LLM to rephrase each instruction without changing the underlying meaning. See the specific meta-prompt in Appendix \ref{sec:appendixB}, Listing \ref{lst:prompt_rephrase}. 
        \item \textbf{Permute}: Finally, we take $N_{permute}$ instructions and randomly change their order in the current list of instructions.
    \end{itemize}
    \item After expanding the pool using the above three operations, we obtain three candidate sets -- each set being a list of instructions. We rank them by their performance on the validation set and add the best $B$ to the pool. To control for divergence from prior iterations, we additionally introduce a word-level Levenshtein edit distance penalty on the prompts.
    
\end{enumerate}

%% file: latex/experiment_setup.tex
\section{Experimental Setup}
\label{sec:experiment_setup}

\subsection{Tasks, Datasets and Metrics}
We conduct our experiments on two prominent text revision tasks: GEC and Text Simplification. We use the current standard evaluation sets and evaluation metrics for each task. 

\paragraph{Grammatical Error Correction} is the task of correcting text for spelling and grammatical errors. We report results on the Test split of the W\&I+LOCNESS Corpus from the BEA-2019 GEC Shared Task \cite{bryant-etal-2019-bea}. We refer to this dataset as BEA-2019-Test. We evaluate results using $F_{0.5}$ score measured using ERRANT tool\footnote{\url{https://github.com/chrisjbryant/errant}} launched at CodaLab platform\footnote{\url{https://codalab.lisn.upsaclay.fr/competitions/4057}}. Train and dev datasets are sampled from the BEA-2019-Dev dataset (4384 samples).

\paragraph{Text Simplification} is the task of rewriting text in a simpler form without altering its original meaning \cite{saggion2017automatic}. We report results on the ASSET-Test dataset (359 samples) \cite{alva-manchego-etal-2020-asset}  as the main evaluation set. We evaluate results
using the SARI score \cite{xu-etal-2016-optimizing} measured using the EASSE package\footnote{\url{https://github.com/feralvam/easse}} \cite{alva-manchego-etal-2019-easse}. Train and dev datasets are sampled from the ASSET-Dev dataset (2000 samples).




\subsection{Baselines}

\paragraph{Copy} We consider a simple baseline that copies the input text to the output.

\paragraph{Best reference} As a best-case baseline, we provide the scores obtained by the best-performing reference if available.

\vspace{-4pt}

\paragraph{SFT} We consider state-of-the-art Supervised Fine-Tuning (SFT) methods as an alternative to prompt-based learning.

\vspace{-4pt}




\paragraph{Zero Shot} We consider a simple 0-shot prompt, which describes the task as an instruction.

\vspace{-4pt}

\paragraph{Few Shot} We augment the prompt used in the 0-shot setting with a few randomly selected examples demonstrating the task. 

\subsection{APIO Setup}
In addition to evaluating our full proposed method, we also perform an ablation where we only perform the first step of \ourmethod -- namely automatic prompt induction. We denote that in our experiments with \ourmethod-\textsc{Induction-Only}.

\paragraph{Induced prompts:} The induced prompts are derived by extracting three instructions from three randomly selected input-output pairs in the training dataset. To identify the best induced prompt, we perform 10 trials on the validation dataset. 

\paragraph{Optimized prompts:} We optimize the prompts induced in the previous step by continuously adding new instructions using the \textit{Improve} meta-prompt, rephrasing them using the \textit{Rephrase} meta-prompt, and adjusting their order using the \textit{Permute} operation. In our experiments, number of epochs $N_{epochs}$ = 15, $N_{permute}$ = 2, beam size $B$ = 32.

The above parameters were an expedient choice and we did not extensively tune them. With regards to the choice of LLMs used in prompting based approaches, we experiment with two very popular LLMs, namely \texttt{GPT-4o-mini}\footnote{\texttt{gpt-4o-mini-2024-07-18}} and \texttt{GPT-4o}\footnote{\texttt{gpt-4o-2024-05-13}}. We use different generation parameter settings for prompt induction and optimization versus test-time inference. For prompt induction and optimization, we set the temperature $t = 1.0$ and nucleus sampling top-$p$ = $1.0$ for better creativity. For inference, we set temperature $t = 0.0$ and top-$p$ = $0.1$ to decrease randomness in outputs, as instability in outputs leads to worse convergence during  optimization.






%% file: latex/results.tex
\section{Results}


\input{tables/table_ts_gec_merged}
\paragraph{GEC} \ourmethod shows substantial gains over zero-shot, few-shot, and induction-only approaches on GEC (Table \ref{table:ts_gec_merged}). With GPT-4o, \ourmethod achieves an $F_{0.5}$ score of 59.40 (using 10 instructions), which is comparable to the state-of-the-art performance among prompt-based LLMs (which was 57.41 by GPT-3). However, we also note that \ourmethod performance, still falls significantly short of non-prompting SFT ensemble techniques (which scored 72.80), highlighting limitations of solely prompting-based approaches on this task.

\paragraph{Text Simplification} \ourmethod shows significant improvements over baseline methods with both LLMs for the task (Table \ref{table:ts_gec_merged}). Notably, \ourmethod using GPT-4o achieves a SARI score of 49.47, surpassing the previous state-of-the-art score (47.94) for prompt-based methods on the ASSET-Test dataset. 

Overall, we observe that \ourmethod is a highly effective method for automating prompt engineering in text revision tasks. Its strength lies in significantly boosting performance over standard prompting techniques and achieving state-of-the-art for text revision tasks among prompting-based methods—without the need for manual prompt design. The prompt optimization step was shown to be particularly crucial, yielding substantial performance gains, especially in GEC (compare \ourmethod with \ourmethod-\textsc{Induction-Only}). While limitations exist compared to non-prompting methods in GEC, \ourmethod represents a valuable advancement in making prompt engineering easier and accessible. 

%% file: tables/table_ts_gec_merged.tex
\begin{table*}[h]
\small
\centering
\begin{tabular}{c >{\raggedright\arraybackslash}p{0.45\textwidth} l c}
\toprule
\textbf{Task} & \textbf{Approach} & \textbf{LLM} & \textbf{Test Score} \\
\midrule

\multirow{13}{*}{GEC \textsuperscript{*}} & Copy  & -- & 0.00 \\
     & SFT \cite{omelianchuk-etal-2024-pillars} & Multiple & \textbf{72.80} \\     
\cmidrule{2-4}
     & Zero-shot \cite{loem-etal-2023-exploring} & GPT-3 & 53.07 \\
     & Few-shot (16 examples) \cite{loem-etal-2023-exploring} & GPT-3 & \textbf{57.41} \\
     & Few-shot (4 examples) \cite{tang-etal-2024-ungrammatical} & GPT-3.5-Turbo & 53.20 \\
\cmidrule{2-4}
     & Zero-shot (adapted from \cite{loem-etal-2023-exploring}) & GPT-4o-mini & 49.90 \\
     & Few-shot (3 randomly sampled examples) & GPT-4o-mini & 53.01 \\
     & \ourmethodnosp-\textsc{Induction-Only} (3 instructions) & GPT-4o-mini & 38.72 \\
     & \ourmethod (7 instructions) & GPT-4o-mini & \textbf{57.07} \\
\cmidrule{2-4}
     & Zero-shot (adapted from \cite{loem-etal-2023-exploring}) & GPT-4o & 54.66 \\
     & Few-shot (3 examples, randomly sampled) & GPT-4o & 44.50 \\
     & \ourmethodnosp-\textsc{Induction-Only} (3 instructions) & GPT-4o & 43.37 \\
     & \ourmethod (10 instructions) & GPT-4o & \textbf{59.40} \\

\toprule
\multirow{14}{*}{Text Simplification} & Copy  & -- & 20.70 \\     
     & SFT \cite{sheang-saggion-2021-controllable} & T5-base & 45.04 \\  
     & Best reference (ref-0) & -- & \textbf{52.62} \\
\cmidrule{2-4}
\cmidrule{2-4}
     & Few-shot (15 SARI-selected examples, random ordering) \cite{vadlamannati-sahin-2023-metric} & GPT-3-175B & 47.94 \\
\cmidrule{2-4}
     & Zero-shot (adapted from \cite{raheja-etal-2023-coedit}) & GPT-4o-mini & 48.03 \\
     & Few-shot (3 randomly sampled examples) & GPT-4o-mini & 47.16 \\
     & \ourmethod-\textsc{Induction-Only} (3 instructions) & GPT-4o-mini & 48.79 \\
     & \ourmethod (6 instructions) & GPT-4o-mini & \textbf{49.27} \\
\cmidrule{2-4}
     & Zero-shot (adapted from \cite{raheja-etal-2023-coedit}) & GPT-4o & 47.73 \\
     & Few-shot (3 examples, randomly sampled) & GPT-4o & 47.87 \\
     & \ourmethod-\textsc{Induction-Only} (3 instructions) & GPT-4o & 48.93 \\
     & \ourmethod (10 instructions) & GPT-4o & \textbf{49.47} \\
\bottomrule

\end{tabular}
\caption{GEC (BEA-2019-Test | $F_{0.5}$) and Text Simplification results (ASSET-Test | SARI). Results are grouped by baselines (Copy, Best-reference, and SFT), and by other prompt-based methods from different models. \textsuperscript{*}Best reference baseline is unavailable for the GEC task because the BEA-2019-Test dataset has not been published.\vspace{-8pt}}
\label{table:ts_gec_merged} 
\end{table*}



%% file: latex/related_work.tex
\section{Related Work} 

\subsection{LLM Prompting for Text Revision}
\citet{fang2023chatgpt} was the first work to evaluate zero-shot performance using LLMs (ChatGPT in their case) for GEC at both sentence and document levels, finding that ChatGPT exhibited high fluency and produced corrections that enhanced the original text beyond the provided references. 
However, ChatGPT faced challenges in adhering to specific step-by-step formats when given simple prompt instructions. 
More recently, numerous works \cite{coyne2023analyzingperformancegpt35gpt4, loem-etal-2023-exploring, davis-etal-2024-prompting, kaneko-okazaki-2024-controlled, katinskaia-yangarber-2024-gpt} have evaluated both open-source and commercial LLMs on multiple GEC benchmarks, finding that LLMs do not consistently outperform supervised models, especially on minimal edit tasks, and often struggle to balance fluency improvements and preservation of the original meaning. Similarly, many recent works 
\cite{kew-etal-2023-bless, qiang2025redefiningsimplicitybenchmarkinglarge, farajidizaji2024possiblemodifytexttarget} have explored and demonstrated the effectiveness of prompt-based methods for text simplification. 


\subsection{LLM-based Automatic Prompt Optimization (APO)}

Prior work show that LLMs are highly sensitive to seemingly minor prompt variations, such as task specification, information ordering, or stylistic formatting, which can lead to significant performance differences, making prompt engineering a tedious trial-and-error process \cite{li2025surveyautomaticpromptengineering}. 

Several methods have been proposed to automatically identify better-performing prompts, using both continuous and discrete prompt optimization methods \cite{li-liang-2021-prefix, prasad-etal-2023-grips, deng-etal-2022-rlprompt, DBLP:conf/iclr/Zhang0ZSG23}.

Recent work has focused on incorporating LLMs into the optimization process, leveraging their ability to generate natural text. By providing example data to the LLM, \citet{honovich-etal-2023-instruction} generated task instructions directly without an initial prompt. LLMs have also been used to conduct Monte Carlo search \cite{DBLP:conf/iclr/ZhouMHPPCB23} generating additional prompt candidates. Various iterative workflows have been designed to prompt LLMs to self-reflect, analyzing errors and improving upon a previous prompt \cite{pryzant-etal-2023-automatic, ye-etal-2024-prompt}. Evolutionary algorithms \cite{guo2024connecting} suggest systematically refining prompt candidates. 

Our work extends this literature by adapting APO specifically for text revision, combining advances in APO with the unique requirements of text editing tasks.

%% file: latex/appendix_a.tex
\section{\ourmethod Prompt Structure}
\label{sec:appendixA}

\vspace{0.5\baselineskip}
\noindent
\begin{minipage}{\textwidth}
\begin{lstlisting}[style=metapromptstyle, caption={Structure of our proposed prompt, represented as a list of $N$ instructions.}, label={lst:prompt_structure}]
<prompt-header>
* instruction-1
* instruction-2
...
* instruction-N
<prompt-footer>
\end{lstlisting}
\end{minipage}


%% file: latex/appendix_b.tex
\section{Meta-prompts for \ourmethod}
\label{sec:appendixB}


\vspace{0.5\baselineskip}
\noindent
\begin{minipage}{\textwidth}
\begin{lstlisting}[style=metapromptstyle, caption={Meta-prompt for prompt induction for the Text  Simplification task.}, label={lst:prompt_instruction_induction}]
Below is an example of an input-output pair for the Text  Simplification task.

Complex sentence: {input_text}
Simple sentence: {output_text}

You are the prompt engineer. Could you give an instruction for this example?  Do not 
mention any part of the considered texts.
\end{lstlisting}

\begin{lstlisting}[style=metapromptstyle, caption={Meta-prompt for prompts' improvement. It generates new instruction to be added to the existing list. In our setting, the number of instructions in the list varies from 3 to 10.}, label={lst:prompt_improve}]
You are a super-talented prompt engineer. You are working on improvement of the Text  Simplification System

The System has these Instructions:
* {instruction1}
* {instruction2}
* {instruction3}

Below are the examples of System's work:
Input 1: {input_text_1}
System\'s Output 1: {output_text_1}
Gold Output 1: {gold_output_text_1}
Error 1 between System\'s Output 1 and Gold Output 1 for given Input 1: {num1} 
different words.

Input 2: {input_text_2}
System\'s Output 2: {output_text_2}
Gold Output 2: {gold_output_text_2}
Error 2 between System\'s Output 2 and Gold Output 2 for given Input 2: {num2} different words.

Mean error for examples 1-2: 
{ave_num} words.
        
Suggest new instruction to augment existing instructions forcing the System's 
Outputs to be exactly the same as Gold Outputs for the given System's Inputs. You 
need to minimize Errors between System's Outputs and Gold Outputs. Put new 
instruction between <new_instruction> and </new_instruction> tags. Do not use no 
more than two sentences. Do not mention Gold Output. Do not use "newline" symbols in 
your answer. Prioritize fixing cases which have larger error (which have more 
different words).
\end{lstlisting}

\begin{lstlisting}[style=metapromptstyle, caption={Meta-prompt for prompt rephrasing.}, label={lst:prompt_rephrase}]
Generate a variation of the following instruction while keeping the semantic 
meaning, updated instruction must be no more than two sentences

Instruction:{instruction}
Updated instruction:
\end{lstlisting}
\end{minipage}

%% file: latex/appendix_c.tex

\pagebreak
\twocolumn[{}]

\section{Examples of \ourmethod prompts (Prompt Induction only)}
\label{sec:appendixC}

\vspace*{\fill}
\noindent

\begin{minipage}{\textwidth}
\begin{lstlisting}[style=promptstyle, caption={Induced prompt for Grammatical Error Correction task from Table \ref{table:ts_gec_merged}, 3 instructions, GPT-4o-mini.}, label={lst:prompt_inducted_gec_gpt4omini}]
* Identify and correct the grammatical error in the given sentence to improve 
clarity and accuracy.
* Generate a corrected version of the given sentence by identifying and fixing any 
grammatical errors while maintaining the original meaning.
* Given a sentence with grammatical errors, identify and correct the mistakes to 
produce a grammatically accurate version of the sentence.
Sentence: {input_text}
Corrected sentence:
\end{lstlisting}
\end{minipage}

\begin{minipage}{\textwidth}
\begin{lstlisting}[style=promptstyle, caption={Induced prompt for Grammatical Error Correction task from Table \ref{table:ts_gec_merged}, 3 instructions, GPT-4o.}, label={lst:prompt_inducted_gec_gpt4o}]
* Identify and correct any grammatical errors present in the given sentence.
* Identify and correct any grammatical errors in the given sentence to ensure it is 
grammatically accurate.
* Identify any grammatical errors in the provided sentence and correct them, 
ensuring the sentence is grammatically accurate. If the sentence is already correct, 
leave it unchanged.
Sentence: {input_text}
Corrected sentence:
\end{lstlisting}
\end{minipage}

\begin{minipage}{\textwidth}
\begin{lstlisting}[style=promptstyle, caption={Induced prompt for Text Simplification task from Table \ref{table:ts_gec_merged}, 3 instructions, GPT-4o-mini.}, label={lst:prompt_inducted_ts_gpt4omini}]
* Simplify the complex sentence by rephrasing it into a more straightforward version 
while maintaining the original meaning and key information.
* Break down the complex sentence into simpler, more concise sentences while 
maintaining the original meaning. Ensure clarity and ease of understanding in the 
rephrased sentences.
* Simplify the given complex sentence by breaking it into shorter, clearer sentences 
while maintaining the original meaning. Focus on using straightforward language and 
avoiding any unnecessary jargon or complexity.
Complex sentence: {input_text}
Simple sentence:
\end{lstlisting}
\end{minipage}

\begin{minipage}{\textwidth}
\begin{lstlisting}[style=promptstyle, caption={Induced prompt for Text Simplification task from Table \ref{table:ts_gec_merged}, 3 instructions, GPT-4o.}, label={lst:prompt_inducted_ts_gpt4o}]
* Simplify the given complex sentence by breaking it into shorter, clearer sentences 
while retaining the original meaning. Remove any unnecessary abstract language and 
focus on conveying the core ideas directly.
* Simplify the given complex sentence while retaining the original meaning and key 
information. Use simpler language and structure to make the sentence more accessible 
and easier to understand.
* Rewrite the given complex sentence to make it easier to understand while 
preserving its original meaning.
Complex sentence: {input_text}
Simple sentence:
\end{lstlisting}
\end{minipage}

\vspace*{\fill}

%% file: latex/appendix_d.tex

\section{Examples of \ourmethod prompts (Induced and Optimized)}
\label{sec:appendixD}

\noindent

\vspace{0.5\baselineskip}

\begin{figure*}[t]
\centering

\begin{lstlisting}[style=promptstyle, caption={Optimized prompt for Grammatical Error Correction task from Table \ref{table:ts_gec_merged}, 7 instructions, GPT-4o-mini.}, label={lst:prompt_optimized_gec_gpt4omini}]
* Given a sentence with grammatical errors, identify and correct the mistakes to 
produce a grammatically accurate version of the sentence.
* Ensure that the output replicates the phrasing, structure, and punctuation of the 
input exactly, with the primary goal of achieving a completely identical output. 
Prioritize correcting any grammatical errors while maintaining the original meaning 
to minimize discrepancies between the output and the expected format of the input.
* Ensure that the corrected sentence matches the original phrasing, structure, and 
punctuation as closely as possible while correcting grammatical errors, with a 
priority on minimizing the number of differing words. Strive to maintain the 
original meaning in a way that eliminates any discrepancies between 
the output and the expected format of the input.
* Generate a corrected version of the given sentence by identifying and fixing any 
grammatical errors while maintaining the original meaning.
* Make sure the revised sentence closely mirrors the phrasing and structure of the 
original meaning while addressing any grammatical mistakes. Aim for an output that 
is as identical to the reference as possible to ensure accuracy and consistency.
* Identify and correct the grammatical error in the given sentence to improve 
clarity and accuracy.
* Make certain that the output mirrors the original phrasing, structure, and 
punctuation precisely, rectifying grammatical mistakes without introducing any 
discrepancies. Aim for no variations in wording between the output and the original 
format, prioritizing corrections in sentences with more significant errors.
Sentence: {input_text}
Corrected sentence:
\end{lstlisting}
\end{figure*}


\vspace*{\fill}

\begin{figure*}[t]
\centering

\begin{lstlisting}[style=promptstyle, caption={Optimized prompt for Grammatical Error Correction task from Table \ref{table:ts_gec_merged}, 10 instructions, GPT-4o.}, label={lst:prompt_optimized_gec_gpt4o}]
* In cases with higher word differences, carefully review the phrasing and wording 
choices for an exact match with the intended simple form. Ensure the choice of words  
strictly conforms to the simplest possible format while reflecting the input 
structure and content precisely.
* To closely align with the expected output, systematically analyze the original 
sentence and aim for a verbatim transformation using the exact sequence and choice 
of words where simplification allows. Recheck each rewritten sentence to ensure all 
elements of the original are accurately reflected in the simplest possible form, 
prioritizing consistency in language and style.
* Preserve the original sentence structure as closely as possible while simplifying 
the language to reduce word differences significantly. Focus on maintaining the 
specific sequence and choice of words to minimize variation in output.
* Simplify the given complex sentence by breaking down information into shorter, 
clearer sentences and preserving the original meaning.
* Pay close attention to the specific choice of words and phrasing used in the 
original sentences, particularly in cases where there is a large difference in word 
count. Aim to closely match the degree of formality and style while simplifying, 
ensuring the output is concise and directly reflective of the input content.
* Emphasize selecting wording that precisely aligns with the simplest form of the 
input, while significantly reducing word changes by closely mimicking the expected 
output style  and brevity in all cases. Pay particular attention to details that 
show higher word differences, striving to match them exactly.
* Please rewrite the complex sentence in a simpler form, keeping the main idea 
intact so it's easier for everyone to understand.
* Focus on matching the exact vocabulary and phrasing seen in the specific simple 
form associated with each word-for-word transformation to ensure minimal word 
differences. Give priority to adjustments in cases where errors have a substantial 
impact, striving to achieve precision in the chosen wording.
* Concentrate on closely matching the phrasing and style of the original sentence, 
with minimal changes in wording and structure. Aim to reduce word differences, 
especially when errors are significant, to better match the target simplicity and 
tone.
* Rewrite the complex sentence into simpler sentences while preserving the original 
meaning and information.
Complex sentence: {input_text}
Simple sentence:
\end{lstlisting}
\end{figure*}

\begin{figure*}[t]
\centering

\begin{lstlisting}[style=promptstyle, caption={Optimized prompt for Text Simplification task from Table \ref{table:ts_gec_merged}, 6 instructions, GPT-4o-mini.}, label={lst:prompt_optimized_ts_gpt4omini}]
* Make sure the output closely resembles the wording and structure of the provided 
sentence, using straightforward language to maintain its original meaning. Aim to 
keep the same key terms and phrases to limit any variations.
* Ensure that the simplified sentences closely match the structure and wording of 
the simplest version while maintaining the core meaning. Strive to keep changes 
minimal, avoiding significant alterations to the original sentence's intent.
* Split the long sentence into shorter, more straightforward sentences that 
emphasize the key points without unnecessary details. Use plain language to enhance 
clarity.
* Make sure the output closely aligns with the simplest version of the input 
sentence by retaining key terms and phrasing to minimize differences. Prioritize 
preserving the original meaning while following the structure and language of the 
simplest version.
* Simplify the complex sentence by dividing it into shorter, more straightforward 
sentences that preserve the key concepts and important details for better 
comprehension.
* Simplify the complex sentence into a more straightforward and understandable 
version while maintaining its core meaning.

Complex sentence: {input_text}
Simple sentence:
\end{lstlisting}
\end{figure*}

\begin{figure*}[t]
\centering

\begin{lstlisting}[style=promptstyle, caption={Optimized prompt for Text Simplification task from Table \ref{table:ts_gec_merged}, 10 instructions, GPT-4o.}, label={lst:prompt_optimized_ts_gpt4o}]
* In cases with higher word differences, carefully review the phrasing and wording 
choices for an exact match with the intended simple form. Ensure the choice of words 
strictly conforms to the simplest possible format while reflecting the input 
structure and content precisely.
* To closely align with the expected output, systematically analyze the original 
sentence and aim for a verbatim transformation using the exact sequence and choice 
of words where simplification allows. Recheck each rewritten sentence to ensure all 
elements of the original are accurately reflected in the simplest possible form, 
prioritizing consistency in language and style.
* Preserve the original sentence structure as closely as possible while simplifying 
the language to reduce word differences significantly. Focus on maintaining the 
specific sequence and choice of words to minimize variation in output.
* Simplify the given complex sentence by breaking down information into shorter, 
clearer sentences and preserving the original meaning.
* Pay close attention to the specific choice of words and phrasing used in the 
original sentences, particularly in cases where there is a large difference in word 
count. Aim to closely match the degree of formality and style while simplifying, 
ensuring the output is concise and directly reflective of the input content.
* Emphasize selecting wording that precisely aligns with the simplest form of the 
input, while significantly reducing word changes by closely mimicking the expected 
output style and brevity in all cases. Pay particular attention to details that show 
higher word differences, striving to match them exactly.
* Please rewrite the complex sentence in a simpler form, keeping the main idea 
intact so it's easier for everyone to understand.
* Focus on matching the exact vocabulary and phrasing seen in the specific simple 
form associated with each word-for-word transformation to ensure minimal word 
differences. Give priority to adjustments in cases where errors have a substantial 
impact, striving to achieve precision in the chosen wording.
* Concentrate on closely matching the phrasing and style of the original sentence, 
with minimal changes in wording and structure. Aim to reduce word differences, 
especially when errors are significant, to better match the target simplicity and 
tone.
* Rewrite the complex sentence into simpler sentences while preserving the original 
meaning and information.
Complex sentence: {input_text}
Simple sentence:
\end{lstlisting}
\end{figure*}